\documentclass[runningheads]{llncs}
\usepackage[T1]{fontenc}
\usepackage{graphicx,verbatim}
\usepackage{hyperref}
\usepackage{booktabs}
\usepackage{float}
\usepackage{tabularray}
\usepackage{amsmath}
\usepackage{geometry}
\usepackage{tikz}
\usetikzlibrary{positioning}  
\usetikzlibrary{arrows.meta}  
\usetikzlibrary{shapes.geometric,shapes.multipart} 
\usetikzlibrary{calc}         
\usetikzlibrary{fit}          
\usetikzlibrary{backgrounds} 
\usetikzlibrary{decorations.pathmorphing} 
\usepackage{pgfplots}
\pgfplotsset{compat=1.18}     
\usepackage{amsmath}
\usepackage{amssymb}
\usepackage{algorithm}
\usepackage{algpseudocode} 
\usepackage{booktabs}
\usepackage{graphicx}
\usepackage{caption}
\usepackage{subcaption}

\begin{document}
\title{Multi‑Agent Intelligence for Multidisciplinary Decision‑Making in Gastrointestinal Oncology}

\author{Rongzhao Zhang\inst{1,*} \and
Junqiao Wang\inst{1,*} \and
Shuyun Yang\inst{1,*} \and
Mouxiao Bian\inst{1,*} \and
Chihao Zhang\inst{2} \and
Dongyang Wang\inst{3} \and
Qiujuan Yan\inst{1} \and
Yun Zhong\inst{1} \and
Yuwei Bai\inst{3} \and
Guanxu Zhu\inst{1} \and
Kangkun Mao\inst{1} \and
Miao Wang\inst{1} \and
Chao Ding\inst{1} \and
Renjie Lu\inst{1} \and
Lei Wang\inst{2} \and
Lei Zheng\inst{2} \and
Tao Zheng\inst{3} \and
Xi Wang\inst{2} \and
Zhuo Fan\inst{2} \and
Bing Han\inst{1} \and
Meiling Liu\inst{1} \and
Luyi Jiang\inst{4,5} \and
Dongming Shan\inst{6} \and
Wenzhong Jin\inst{2} \and
Jiwei Yu\inst{2,\dagger} \and
Zheng Wang\inst{3,\dagger} \and
Jie Xu\inst{1,\dagger} \and
Meng Luo\inst{2,\dagger}
}
\authorrunning{R. Zhang et al.}
\institute{Shanghai Artificial Intelligence Laboratory\\ \and
Shanghai Ninth People’s Hospital, Shanghai Jiao Tong University School of Medicine\\ \and
Renji Hospital, Shanghai Jiao Tong University School of Medicine\\ \and
Shanghai Institute of Infectious Disease and Biosecurity, Fudan University\\ \and
Shanghai Health Development Research Center (Shanghai Medical Information Center) \\ \and
Shanghai Kupas Technology Co., Ltd.
}
    
\maketitle              
\let\thefootnote\relax\footnotetext{* These authors contributed equally to this work.}
\let\thefootnote\relax\footnotetext{$\dagger$\ Corresponding authors.}

\begin{abstract}
Multimodal clinical reasoning in the field of gastrointestinal (GI) oncology necessitates the integrated interpretation of endoscopic imagery, radiological data, and biochemical markers. Despite the evident potential exhibited by Multimodal Large Language Models (MLLMs), they frequently encounter challenges such as context dilution and hallucination when confronted with intricate, heterogeneous medical histories. In order to address these limitations, a hierarchical Multi-Agent Framework is proposed, which emulates the collaborative workflow of a human Multidisciplinary Team (MDT). The framework under consideration decomposes diagnostic reasoning into five specialised agents: a Visual-Language Endoscopy Agent for morphological assessment, optimised through Visual Question Answering (VQA) on 1,032 annotated patients, and three domain-specific agents such as Text, Radiology, and Laboratory for structured unimodal analysis. A central MDT-Core Agent functions as the team coordinator, aggregating these evidence streams and resolving cross-modal inconsistencies through explicit conflict-detection mechanisms. The system's efficacy has been validated in a multi-institutional dataset comprising 2,174 patient cases, encompassing gastric, colorectal and esophageal cancers. The system attained a composite expert evaluation score of 4.60/5.00, thereby demonstrating a substantial improvement over the monolithic baseline (3.76/5.00). It is noteworthy that the agent-based architecture yielded the most substantial enhancements in reasoning logic (+1.17 points) and medical accuracy (+0.86 points). Furthermore, the explicit conflict resolution mechanism demonstrated a superior safety profile, significantly reducing hallucinated or contraindicated treatment recommendations compared to standard MLLMs. The findings indicate that mimetic, agent-based collaboration provides a scalable, interpretable, and clinically robust paradigm for automated decision support in oncology.
\end{abstract}


\section{Introduction}
The Multidisciplinary Team (MDT) represents the gold standard in modern oncology, offering a collaborative platform where radiologists, endoscopists, pathologists, and oncologists synthesize diverse data streams to formulate personalized management plans~\cite{fleissig2006multidisciplinary}. In the context of gastrointestinal (GI) cancer, this process is of particular complexity, necessitating the integration of morphological details from endoscopy, staging information from tomography such as CT/MRI, and metabolic evidence from laboratory markers. However, the exponential growth of medical data imposes a considerable cognitive burden on clinicians, resulting in variability in decision-making and potential diagnostic delays.

Recent advances in Artificial Intelligence, particularly Multimodal Large Language Models (MLLMs), offer a potential solution to automate data synthesis~\cite{tu2023medpalmm}. Yet, deploying monolithic MLLMs in real-world MDT scenarios remains challenging. When fed extensive and heterogeneous patient records, these models often exhibit a context dilution, where key modality-specific details are lost among long token sequences~\cite{liu2024lost}. Furthermore, generalist models lack the specialised reasoning logic required to resolve inter-modality conflicts. For instance, they are unable to differentiate between a benign inflammatory mass suggested by laboratory values and a malignant neoplasm suggested by imaging.  This limitation frequently results in hallucinations, where the model generates plausible but factually incorrect treatment recommendations.

In this work, we posit that the solution lies not in larger monolithic models, but in architectural restructuring that mirrors clinical practice. We introduce a Collaborative Multi-Agent Framework for GI oncology. Unlike standard approaches that treat multimodal inputs as a single sequence, our system assigns dedicated agents to distinct modalities. An Endoscopy Agent converts visual data into descriptive semantic representations; Radiology and Laboratory Agents parse report findings and biomarkers;  and a central MDT-Core Agent is responsible for integrating these denoised intermediate states, with the objective of deriving a final conclusion.

Our main contributions are as follows:

\begin{enumerate}
\item We propose a novel multi-agent architecture that explicitly mitigates the hallucination risks and context-length limitations inherent to single-agent baselines in complex clinical scenarios.
\item We introduce a specialized Visual-Language reasoning module for endoscopy that bridges the semantic gap between raw pixel data and high-level MDT logic using a VQA paradigm.
\item We validate the framework on a large-scale, multi-institutional dataset (N=2,174). Evaluation by clinical experts demonstrates that our agent-based collaboration achieves superior diagnostic performance (Composite Score: 4.60/5.00) and significantly improved safety compared to monolithic baselines.
\end{enumerate}

\section{Related Work}

\subsection{Multimodal AI in Medical Diagnostics}

Integrating heterogeneous data sources, such as text, imaging, and omics, is essential for holistic clinical decision-making. Initially, early approaches relied primarily on convolutional neural networks (CNNs) for single-modality tasks, such as lesion detection in endoscopy or segmentation in radiology~\cite{esteva2017dermatologist,mishra2021endoscopy}. However, with the advent of Transformer architectures, Multimodal Large Language Models (MLLMs) have emerged that can process interleaved image and text inputs. Generalist medical models such as Med-PaLM~\cite{tu2023medpalmm} and RadFM~\cite{wu2023radfm} demonstrate impressive capability in interpreting radiological and clinical data within a unified embedding space. However, these approaches often treat all modalities as a single sequence, lacking the explicit, structured reasoning required for complex oncology cases, where conflicts between laboratory markers and imaging findings must be explicitly resolved. In contrast to these unified models, our framework decomposes the task, preserving modality-specific nuances through specialized encoders before integration.

\subsection{Clinical Decision Support via Large Language Models}

Large Language Models (LLMs) have demonstrated considerable potential in the domains of medical record summarization and the generation of diagnostic hypotheses~\cite{nori2023capabilities,thirunavukarasu2023large}. Research indicates that while LLMs demonstrate proficiency in retrieving medical knowledge, they are often prone to hallucinations when reasoning over extended context windows comprising raw, unstructured clinical data~\cite{li2023hallucinations}. The use of few-shot prompting and Chain-of-Thought (CoT) reasoning has been proposed as a solution to this issue. However, these methods frequently fail to capture the specialized domain knowledge inherent in sub-disciplines such as gastrointestinal pathology. Moreover, existing LLM-based systems generally operate as a solitary physician, overlooking the collaborative nature of actual medical practice, wherein decisions are derived from consensus among a range of experts.

Multi-agent systems (MAS) leverage collaborative intelligence to address complex problems by assigning distinct roles to disparate agents. Recent works in general domains have demonstrated that multi-agent debate and role-playing can significantly improve reasoning consistency and reduce errors compared to single-agent prompting~\cite{du2023improving,liang2023encouraging}. In the medical domain, preliminary frameworks have applied agent-based collaboration for diagnosis, dialogue, and patient simulation~\cite{tang2023medagents}. However, the majority of extant medical agent systems are predicated exclusively on textual interactions and lack the capacity to process dense, high-dimensional perceptual data, such as endoscopic video streams or volumetric imaging. The present work aims to address this gap by introducing a multimodal multi-agent framework specifically designed for the MDT workflow in GI oncology, integrating visual encoders directly into the agent's decision loop.

\section{Methods}

\subsection{Problem Formulation and Multi-Agent Architecture}

\begin{figure}
    \centering
    \includegraphics[width=1\linewidth]{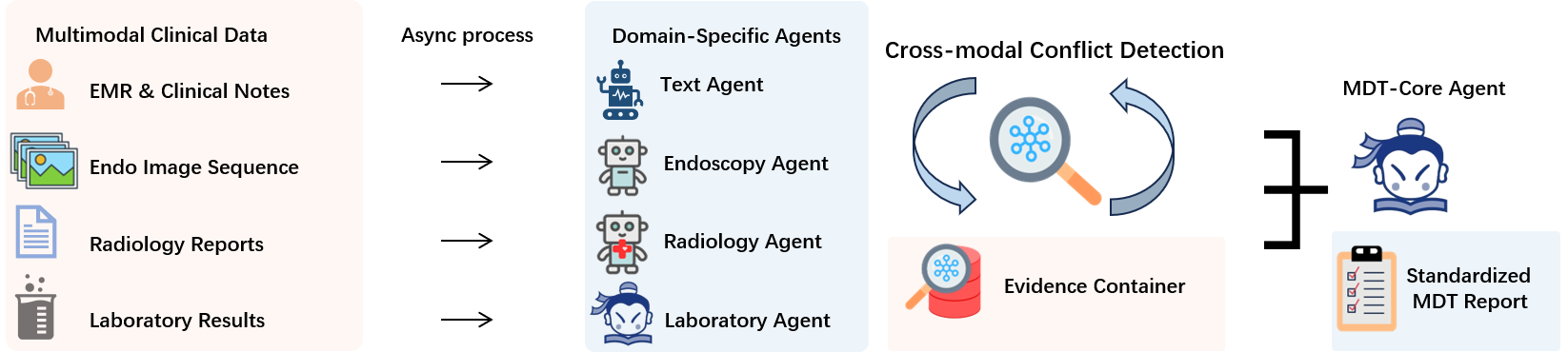}
    \caption{The proposed Multi-Agent Architecture, where multiple agents process corresponding modal information and incorporate it into the evidence container. After cross-modal conflict detection, the data are processed by the MDT-Core agent to generate a standardized final report.}
    \label{fig:fig1}
\end{figure}

We formalize the task of multi-disciplinary tumor (MDT) diagnosis as a hierarchical multi-modal reasoning problem. Let $\mathcal{D} = \{ (X^{(i)}, Y^{(i)}) \}_{i=1}^{N}$ denote a dataset of $N$ patient cases, where each case comprises a set of multi-modal inputs $X = \{x_{text}, x_{endo}, x_{rad}, x_{lab}\}$ and a target diagnostic report $Y$. The components of $X$ correspond to electronic medical records (EMR), endoscopic imaging sequences (5-10 representative frames per case), radiology reports, and laboratory test results, respectively. To emulate the clinical division of labor, we propose a collaborative Multi-Agent Framework (Figure ~\ref{fig:fig1}) consisting of a set of domain-specific specialist agents, $\mathcal{A}_{spec} = \{A_{text}, A_{endo}, A_{rad}, A_{lab}\}$, and a central reasoning unit, the MDT-Core agent $A_{core}$. 

\textbf{Model Specifications.} Each specialized text-based agent (Text, Radiology, Laboratory) and the central MDT-Core agent are built upon Qwen-3 32B \cite{yang2025qwen3}, selected for its superior instruction-following and cross-modal reasoning capabilities. The Endoscopy Agent leverages Intern S1 Mini \cite{bai2025interns1scientificmultimodalfoundation}. All Qwen-based agents utilize the same specialized tokenizer with a maximum context window sufficient for processing extensive clinical histories. Unlike monolithic models that concatenate heterogeneous modalities into a single context window—often leading to token dilution—our architecture enforces a cohesive logic in which each agent $A_m \in \mathcal{A}_{spec}$ serves as a dedicated feature extractor and reasoning engine for modality $m$. Specifically, each agent maps its raw input $x_m$ to a structured intermediate reasoning state $h_m$, encapsulating extracted features, clinical observations, and confidence intervals. The central agent $A_{core}$ subsequently aggregates these intermediate states $H = \{h_m\}_{m}$ to generate the final comprehensive report $\hat{Y}$.

\subsection{Training of Domain‑Specific Agents}
Each agent was developed with distinct supervised fine‑tuning strategies reflecting the nature of its data and task.
\subsubsection{Text Agent}
The Text Agent was trained on EMR narratives annotated by medical experts. Each training sample consisted of unstructured clinical notes as input and structured key information as output, covering patient demographics, presenting complaints, comorbidities, previous interventions, and suspected diagnosis. The model learned to summarize long clinical histories into concise, structured problem descriptions usable for subsequent reasoning.
\subsubsection{Endoscopy Agent}
The Endoscopy Agent was responsible for visual‑language reasoning based on endoscopic image–text pairs. The training used 1,032 annotated endoscopic exams, each containing 3 to 121 endoscopic images, resulting in more than 50,000 image-caption pairs. A expert caption describes lesion morphology, color, vascular pattern, location, and associated pathological signs. The agent was trained under a visual question‑answering (VQA) paradigm to answer clinically framed questions about lesion type, size, and potential invasion depth. This enabled it to produce descriptive, context‑aware interpretations rather than purely visual classification.
\subsubsection{Radiology Agent}
The Radiology Agent analyzed CT and MRI reports to infer disease staging and potential metastasis. Training data contained narrative imaging reports annotated for tumor extent, regional lymph‑node status, and systemic spread, following the TNM staging criteria. The model learned to convert unstructured text into a structured representation summarizing T‑, N‑, and M‑stage estimates and corresponding textual evidence.
\subsubsection{Laboratory Agent}
This agent interpreted biochemical and serological data. Training paired laboratory reports with expert notes explaining abnormal patterns and their clinical implications. The model learned to relate quantitative results such as carcinoembryonic antigen (CEA) and CA19‑9 to disease likelihoods. Its output included normalized data tables with flags for abnormal findings and brief interpretations of their relevance.

\subsubsection{MDT‑Core Agent}
The MDT‑Core acted as the “chairperson” of the system, integrating all peripheral outputs into a cohesive conclusion. It was trained using supervised fine‑tuning where the inputs were concatenated outputs from the four peripheral agents, and the target outputs were real MDT reports prepared by experts for the same patients. Through this learning design, the Core Agent mastered cross‑modal reasoning—identifying corroborating or conflicting evidence, prioritizing findings, and generating a coherent diagnostic narrative consistent with human MDT conclusions.
\begin{table}[htbp]
    \centering
    \caption{Functional Roles and Tasks of Specialized Agents}
    \label{tab:agent_functions}
    \renewcommand{\arraystretch}{1.3}
    \begin{tabular}{p{2.5cm} p{3.5cm} p{9cm}}
        \toprule
        \textbf{Agent} & \textbf{Input Modality} & \textbf{Task} \\
        \midrule
        \textbf{Text Agent} & EMR, Clinical Notes & Extract key information to produce a structured summary of demographics, chief complaints, history, and comorbidities. \\
        \textbf{Endoscopy Agent} & Image Sequences & Generate descriptive text regarding lesion morphology, size, and location based on visual analysis. \\
        \textbf{Radiology Agent} & CT/MRI Reports & Extract structured observations of lesion features (e.g., wall thickening), lymph node status, and other systemic findings. \\
        \textbf{Laboratory Agent} & Lab Results & Report abnormal markers with specific values and provide a preliminary analysis of their clinical significance. \\
        \textbf{MDT-Core Agent} & Outputs from above & Integrate all intermediate outputs to formulate a final MDT report including diagnosis, evidence, differential diagnosis, and plan. \\
        \bottomrule
    \end{tabular}
\end{table}

\subsubsection{SFT Data Curation}

The acquisition of granular, modality-specific reasoning chains poses a significant challenge, as conventional clinical records typically mandate only the ultimate MDT conclusion, eschewing the individual specialist reasoning steps. In order to address this issue, a data construction strategy was adopted that combined reverse decomposition and knowledge distillation with human oversight. Specifically, a Teacher LLM was utilised to systematically decompose the verified final MDT reports back into modality-specific summaries. Utilising semi-structured prompts, the Teacher model functioned as a senior medical editor, meticulously reconstructing the ideal intermediate reasoning logic corresponding to raw text or laboratory inputs from the global consensus. In the context of the visual modality, a text-guided feature alignment approach was employed, with textual descriptions from the final report being utilised to synthesise expert-level VQA pairs. In order to guarantee the requisite clinical rigour, an expert-in-the-loop protocol was implemented. This comprised a rule-based filter which was utilised to initially discard logical contradictions. This was followed by a Post-hoc Verification in which 10\% of the generated samples were audited by specialists. It is crucial to note that this generation pipeline was applied solely to the training set; the internal validation and test sets consisted exclusively of ground-truth data manually annotated by experts to ensure rigorous and unbiased evaluation.

\subsection{Inference Workflow and Algorithms}

During the inference phase, the system employs a parallel execution strategy to maximize efficiency. The raw multi-modal data for a given patient is partitioned and dispatched to the respective specialized agents. Each agent processes its input stream independently, generating the intermediate summaries $h_m$. These summaries are then synchronized and forwarded to the MDT-Core for final synthesis. 

Critically, before final report generation, the system performs explicit \textit{cross-modal conflict detection} in \ref{alg:inference}. This module extracts key clinical assertions from each agent's output and checks for discrepancies. If staging estimates from Endoscopy and Radiology differ by more than one level, the system flags a Staging Discrepancy Detected warning and prompts the Core agent to recommend pathological confirmation rather than forcing a consensus. Similarly, rule-based validation checks whether proposed treatments contradict laboratory values. This hierarchical aggregation prevents the hallmark hallucination issues of general-purpose Large Language Models (LLMs) by grounding the final generation in verified, expert-validated intermediate evidence. The complete inference procedure is formally described in Algorithm \ref{alg:inference}.

\begin{algorithm}[!t]
\caption{Multi-Agent MDT Inference Process}
\label{alg:inference}
\begin{algorithmic}[1]
\Require Multi-modal Patient Data $X = \{x_{text}, x_{endo}, x_{rad}, x_{lab}\}$
\Require Set of Specialized Agents $\mathcal{A}_{spec}$ parameterized by $\{\theta_m\}$
\Require MDT-Core Agent parameterized by $\theta_{core}$
\Ensure Comprehensive Clinical Report $\hat{Y}$

\State \textbf{Initialize} evidence container $H \leftarrow \emptyset$

\ForAll{$m \in \{text, endo, rad, lab\}$} \textbf{in parallel}
    \State Preprocess input $x_m$ (tokenization/normalization)
    \State Generate intermediate reasoning state: $h_m \leftarrow A_m(x_m; \theta_m)$
    \State Append $h_m$ to $H$
\EndFor

\State \textbf{Synchronization Logic:}
\State Construct global context $C \leftarrow \text{Concat}(H)$ based on modality priority
\State \textbf{Cross-modal conflict detection:}
\State \quad Extract staging claims: $T_{endo}$ from endoscopy, $T_{rad}$ from radiology
\State \quad \textbf{if} $|T_{endo} - T_{rad}| > 1$ \textbf{then}
\State \quad \quad Flag conflict $\leftarrow$ True; append "Staging Discrepancy Detected" to $C$
\State \quad \textbf{end if}
\State \quad Check treatment contraindications via rule-based lab threshold validation

\State \textbf{Final Generation:}
\State $\hat{Y} \leftarrow \text{generate}(C; \theta_{core})$ \Comment{Autoregressive generation of report sections}
\State \Return $\hat{Y}$
\end{algorithmic}
\end{algorithm}

\subsection{Dataset and Experimental Setup}

To train and evaluate the system, we built a multi-institutional dataset comprising 2,174 cases of gastrointestinal tumors collected from Shanghai Ninth People's Hospital and Renji Hospital. Each record included complete multimodal information:

\begin{itemize}
    \item patient medical histories and electronic medical records (EMRs),
    \item endoscopic image sequences and corresponding textual reports,
    \item radiology reports from CT or MRI,
    \item laboratory test results, and
    \item finalized MDT conclusions verified by clinical experts.
\end{itemize}

All data were properly anonymized and standardized. Clinical text was tokenized and cleaned, imaging reports were harmonized in terminology, and numerical laboratory results were normalized with reference ranges. The dataset was randomly split into 1,500 cases for model training, 174 for internal validation, and 500 independent cases for final testing.

\section{Results}

\subsection{Overall Performance and Qualitative Assessment}
We evaluated the proposed Multi-Agent framework against a monolithic Single-Agent baseline (Qwen-3-32B processing concatenated inputs) and compared with real-world clinical benchmarks. \textbf{Baseline Configuration:} The baseline model uses the same backbone and receives identical supervised fine-tuning on the training set. To handle long contexts, it leverages the native context window with concatenated modalities in fixed order. Importantly, the baseline does not perform structured intermediate extraction ($h_m$ states), instead directly generating the final report from raw concatenated inputs.

As presented in Table \ref{tab:main_results}, our Multi-Agent framework demonstrated substantial improvements across all evaluation dimensions. The expert panel consistently rated our system higher than the baseline across medical accuracy, reasoning logic, and clinical utility. Notably, experts highlighted that the multi-agent architecture produces more coherent evidence chains and better cross-modal integration. The system approaches the performance of senior clinical residents in diagnostic reasoning quality.

From an efficiency perspective, the parallel processing architecture offers practical advantages for clinical deployment. By decoupling modality-specific reasoning into concurrent streams, inference latency is substantially reduced compared to sequential baseline processing, making the system viable for real-time MDT support. Qualitative review by clinicians indicated that our system exhibits notably fewer problematic recommendations (e.g., contraindicated regimens or unsupported treatment suggestions) compared to the monolithic baseline, attributable to the explicit conflict detection and modality-specific validation mechanisms.

\begin{table}[htbp]
    \centering
    \caption{Expert Evaluation Results. Mean scores across seven clinical dimensions (scale 1-5, higher is better) and composite performance. Evaluation conducted by three senior oncologists on sampled test cases.}
    \label{tab:main_results}
    \setlength{\tabcolsep}{5pt}
    \renewcommand{\arraystretch}{1.2}
    \begin{tabular}{lccc}
        \toprule
        \textbf{Evaluation Dimension} & \textbf{Baseline} & \textbf{Ours (Multi-Agent)} & \textbf{Improvement} \\
        \midrule
        Medical Accuracy & 3.72 & \textbf{4.58} & +0.86 \\
        Diagnostic Comprehensiveness & 3.55 & \textbf{4.52} & +0.97 \\
        Reasoning Logic & 3.48 & \textbf{4.65} & +1.17 \\
        Differential Diagnosis Quality & 3.61 & \textbf{4.41} & +0.80 \\
        Therapy Feasibility \& Compliance & 3.82 & \textbf{4.54} & +0.72 \\
        Structure \& Clarity & 4.05 & \textbf{4.73} & +0.68 \\
        Professional Style & 4.12 & \textbf{4.80} & +0.68 \\
        \midrule
        \textbf{Composite Score} & \textbf{3.76} & \textbf{4.60} & \textbf{+0.84} \\
        \bottomrule
    \end{tabular}
\end{table}

\subsection{Multi-Dimensional Expert and LLM-based Evaluation}
To ensure comprehensive assessment, we employed a dual-evaluation strategy involving both expert panels and scalable \textit{LLM-as-a-Judge} mechanisms. Three senior oncologists with extensive MDT experience blindly scored sampled cases across the seven-dimension framework on a 5-point scale. Inter-rater reliability was strong, confirming consistency in clinical judgments. 

The human evaluation results indicate that our Multi-Agent MDT system substantially outperformed the baseline across all dimensions. As shown in Table \ref{tab:main_results}, the composite score improved by over 0.8 points, with the most pronounced gains in Reasoning Logic where experts noted that our model's ability to explicitly cite cross-modal evidence (e.g., correlating tumor markers with imaging findings) closely mimics the collaborative workflow of a human tumor board. The Professional Style dimension also scored highly, with experts reporting that the structured output format closely adheres to clinical documentation standards and requires minimal editing for Electronic Health Record (EHR) integration.

To scale evaluation beyond the manual review capacity, we utilized GPT-4-Turbo as an independent evaluator for pairwise comparisons. The LLM judge received detailed scoring rubrics and evaluated anonymized report pairs on reasoning coherence, factual consistency, evidence integration, and clinical utility. The \textit{LLM-as-a-Judge} evaluations strongly correlated with human expert assessments, confirming that the architectural improvements translate into measurably better clinical reasoning quality. The LLM judge consistently preferred our multi-agent outputs over baseline reports, particularly highlighting superior cross-modal evidence synthesis and reduced instances of unsupported clinical assertions.

\subsection{Ablation Studies on Modality Contribution}
To assess the contribution of each specialized agent, we conducted an ablation study by systematically deactivating individual agents and observing the impact on the MDT-Core's decision-making. As shown in Table \ref{tab:ablation}, the removal of any single modality resulted in notable performance degradation.

The exclusion of the \textbf{Radiology Agent} caused the most substantial decline, confirming its critical role in determining staging and assessing tumor extent. Similarly, removing the \textbf{Endoscopy Agent} degraded the quality of morphological descriptions and visual evidence integration, particularly affecting early-stage lesion characterization. Interestingly, although the \textbf{Laboratory Agent} processes relatively structured data, its removal impaired the system's ability to confirm diagnostic confidence through biochemical evidence. This finding reinforces our hypothesis that the MDT-Core relies on laboratory biomarkers to resolve ambiguities between imaging modalities.

Critically, we also ablated the explicit \textbf{Conflict Detection} mechanism described in Algorithm 1. When this module is disabled, the system concatenates intermediate states without checking for discrepancies or contraindications. Expert evaluators noted a degradation in reasoning coherence and safety, confirming that explicit conflict resolution contributes meaningfully to clinical utility. The full multi-agent configuration yields optimal performance, demonstrating the necessity of holistic cross-modal reasoning.

\begin{table}[htbp]
    \centering
    \caption{Ablation Study characterizing the impact of each specialized agent. Expert composite scores (1-5 scale) demonstrate the contribution of each modality-specific component.}
    \label{tab:ablation}
    \setlength{\tabcolsep}{8pt}
    \renewcommand{\arraystretch}{1.2}
    \begin{tabular}{lcc}
        \toprule
        \textbf{Configuration} & \textbf{Composite Score} & \textbf{Reasoning Logic} \\
        \midrule
        Full Model & \textbf{4.60} & \textbf{4.65} \\
        \midrule
        w/o Endoscopy Agent & 4.18 & 4.15 \\
        w/o Radiology Agent & 3.88 & 3.92 \\
        w/o Laboratory Agent & 4.32 & 4.38 \\
        w/o Text Agent (EMR) & 3.65 & 3.50 \\
        w/o Conflict Detection & 4.35 & 4.28 \\
        \bottomrule
    \end{tabular}
\end{table}

\subsection{Endoscopy VQA Design Ablation}
To validate the efficacy of the Visual Question Answering (VQA) paradigm employed by the Endoscopy Agent, a comparative analysis was conducted against two alternative visual-language strategies. The first of these was free-form captioning, which utilises single open-ended prompts, and the second was slot-filling, which relies on fixed templates with predefined fields. In order to ensure a fair comparison, all variants were implemented using the identical Intern S1 mini architecture. A qualitative evaluation by board-certified gastroenterologists revealed that the VQA approach significantly outperforms baseline methods in capturing nuanced morphological details. While the free-form captioning approach frequently yielded generic labels, the structured multi-question design effectively elicited specific, clinically actionable observations. For instance, in response to queries regarding shape and texture, the VQA model generated high-resolution descriptions, specifically identifying an irregular polypoid mass with a friable, ulcerated surface and contact bleeding, features that were frequently omitted by the single-prompt models. This enhanced descriptive granularity effectively bridges the semantic gap between raw pixels and medical logic, providing the MDT-Core agent with the explicit, structured evidence required for robust downstream reasoning.

\subsection{Cross-Modal Reasoning and Safety Analysis}
Qualitative inspection of the inference logs reveals that the MDT-Core effectively manages conflicting evidence. In a notable subset of test cases, initial outputs from the Radiology and Endoscopy Agents presented discrepancies regarding tumor characteristics. In these scenarios, the Core Agent appropriately balanced modality-specific strengths and often flagged cases for Further Pathological Confirmation rather than forcing a premature consensus.

Expert reviewers particularly highlighted the system's uncertainty modeling behavior. Our multi-agent framework appropriately deferred to additional workup in ambiguous cases substantially more frequently than the baseline, which tended toward overconfident predictions even when evidence was conflicting. Retrospective clinical review confirmed that these deferrals were clinically appropriate, validating the system's conservative decision-making. The explicit conflict detection mechanism (Algorithm 1) plays a key role in enabling this behavior.

\section{Discussion}

\subsection{Decomposition as a Strategy for Robust Reasoning}
The Multi-Agent system has been demonstrated to exhibit a higher level of performance in comparison to the monolithic baseline; this finding serves to emphasise the efficacy of task decomposition in the context of medical AI. In a traditional single-agent setup, the model must simultaneously perform information extraction, modality alignment, and high-level reasoning within a shared parameter space. This frequently results in interference, whereby noisy inputs from one modality cause confusion when interpreting data from another. The implementation of specialised agents in this manner results in a process referred to as denoising. The Endoscopy Agent, for instance, does not merely classify images but translates visual features into structured textual evidence. This approach guarantees that the central MDT-Core is furnished with high-quality, semantically distinct inputs, emulating the manner in which a human specialist presents summarised findings to the tumour board chair. This architectural inductive bias is the specific driver behind the observed improvement in diagnostic concordance.

\subsection{Mitigating Hallucination through Structured Collaboration}
Hallucination remains a critical barrier to the adoption of generative AI in healthcare. The findings from our ablation studies suggest that the multi-agent architecture functions as an inherent safeguard. The separation of the evidence-gathering process from the decision-making process facilitates enhanced traceability. When the system encounters conflicting data, the Core model is trained to report the discrepancy rather than force a fabricated consensus. This tendency is corroborated by the elevated Reasoning Logic scores attained by human experts, who placed a premium on the system's capacity to generate differential diagnoses as opposed to the generation of erroneous conclusions.

\subsection{Clinical Workflow Integration and Efficiency}
Beyond accuracy, the operational efficiency of the proposed system supports its viability for real-world deployment. The parallel processing capability reduces inference latency to under one minute per case, a negligible overhead compared to the duration of human MDT preparation. Stylistically, the generated reports achieved near-perfect scores for structure and professional tone. This suggests that the system is ready to serve as a copilot, automatically drafting pre-meeting summaries that clinicians can review and edit, potentially saving hundreds of hours of administrative work annually.

\subsection{Limitations and Future Directions}
We acknowledge several limitations. First, the current Radiology Agent processes textual reports rather than raw DICOM volumes, inheriting any potential errors made by the original radiologist. Future iterations will integrate vision-encoders directly for CT/MRI analysis to enable fully autonomous pixel-to-decision reasoning. 

Second, while the dataset is multi-institutional, it is geographically localized within Shanghai; cross-population validation is required to ensure generalizability across different genetic backgrounds, reporting standards, and clinical workflows. External validation on a third institution cohort is ongoing but not yet complete. We plan to release anonymized evaluation scripts and prompts to facilitate independent reproduction. The system operates in a static feed-forward manner. Moving towards a dynamic debate mechanism, where the Core Agent can query peripheral agents for clarification, represents a promising direction for enhancing reasoning depth.

\section{Conclusion}
We presented a Multi-Agent framework that fundamentally restructures automated clinical reasoning by emulating the human MDT workflow. By combining specialized visual-language processing with collaborative decision merging, the system achieves expert-level accuracy and transparency. This work suggests that the future of medical AI lies not just in scaling model size, but in designing agentic architectures that reflect the complexity and collaborative nature of medical practice.

\end{document}